\documentclass[10pt,twocolumn,letterpaper]{article}

\usepackage{cvpr}
\usepackage{times}
\usepackage{epsfig}
\usepackage{xspace}
\usepackage{color}
\usepackage{multirow}
\usepackage{graphics}
\usepackage{adjustbox}
\usepackage{subfigure}
\usepackage{amsmath}
\usepackage[noend]{algpseudocode}
\usepackage{algorithmicx,algorithm}
\usepackage{paralist} 
\usepackage{enumitem}
\usepackage{graphicx} 
\usepackage{epstopdf}
\usepackage{booktabs} 

\usepackage[pagebackref=true,breaklinks=true,letterpaper=true,colorlinks,citecolor=blue,linkcolor=blue,bookmarks=false]{hyperref}

\long\def\ignorethis#1{}

\definecolor{Gray}{rgb}{0.35,0.35,0.35}
\definecolor{Blue}{rgb}{0,0.2,0.8}
\definecolor{Red}{rgb}{0.8,0.2,0}
\definecolor{Green}{rgb}{0.0,0.5,0.1}
\definecolor{Gray}{rgb}{0.4,0.4,0.4}

\newlength\paramargin
\newlength\figmargin
\newlength\secmargin

\usepackage{array}
\newcolumntype{L}[1]{>{\raggedright\let\newline\\\arraybackslash\hspace{0pt}}m{#1}}
\newcolumntype{C}[1]{>{\centering\let\newline\\\arraybackslash\hspace{0pt}}m{#1}}
\newcolumntype{R}[1]{>{\raggedleft\let\newline\\\arraybackslash\hspace{0pt}}m{#1}}

\graphicspath{{figure/eps/}}

\usepackage{times}
\usepackage{epsfig}
\usepackage{graphicx}
\usepackage{amsmath}
\usepackage{amssymb}
\usepackage{booktabs}
\usepackage{subfigure}
\usepackage{multirow}

\graphicspath{ {figures/} }

\usepackage[breaklinks=true,bookmarks=false]{hyperref}

\cvprfinalcopy 

\def\cvprPaperID{632} 

\begin{document}

\title{Learning a Discriminative Feature Network for Semantic Segmentation}

\author{
    Changqian Yu$^1$
    \hspace{10pt}
    Jingbo Wang$^2$
    \hspace{10pt}
    Chao Peng$^3$
    \hspace{10pt}
    Changxin Gao$^1$\thanks{Corresponding author.}
    \hspace{10pt}
    Gang Yu$^3$
    \hspace{10pt}
    Nong Sang$^1$\\
    $^1$Key Laboratory of Ministry of Education for Image Processing and Intelligent Control,\\School of Automation, Huazhong University of Science and Technology\\
    $^2$Key Laboratory of Machine Perception, Peking University\\
    $^3$Megvii Inc. (Face++)\\
{\tt\small \{changqian\_yu,cgao,nsang\}@hust.edu.cn,wangjingbo1219@pku.edu.cn,\{pengchao,yugang\}@megvii.com}
}

\maketitle
\begin{abstract}
   Most existing methods of semantic segmentation still suffer from two aspects of challenges: intra-class inconsistency and inter-class indistinction. To tackle these two problems, we propose a Discriminative Feature Network~(DFN), which contains two sub-networks: Smooth Network and Border Network. Specifically, to handle the intra-class inconsistency problem, we specially design a Smooth Network with Channel Attention Block and global average pooling to select the more discriminative features. Furthermore, we propose a Border Network to make the bilateral features of boundary distinguishable with deep semantic boundary supervision. Based on our proposed DFN, we achieve state-of-the-art performance $\textbf{86.2\%}$ mean IOU on PASCAL VOC 2012 and $\textbf{80.3\%}$ mean IOU on Cityscapes dataset.
\end{abstract}

\section{Introduction}
\label{sec:introduction}

\begin{figure}[t]
\includegraphics[width=\linewidth]{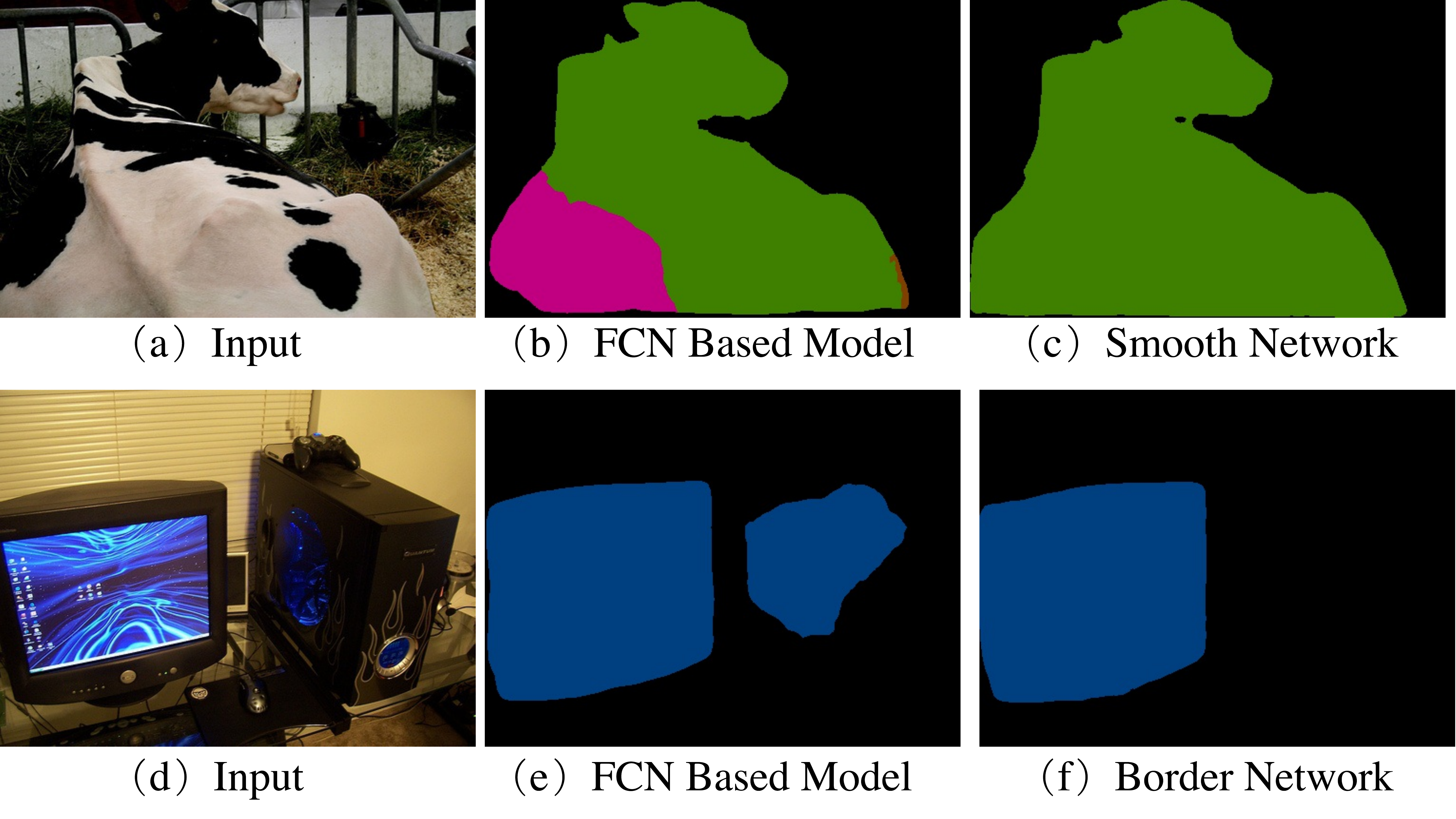} 
\vspace{-4mm}
\caption{Hard examples in semantic segmentation. The second column is the output of FCN based model. The third column is the output of our proposed approach. In the first row, the left bottom corner of the cow is recognized as a horse. This is the \textbf{Intra-class Inconsistency} problem. In the second row, the computer case has the similar blue light and black shell with the computer screen, which is hard to distinguish. This is the \textbf{Inter-class Indistinction} problem.}
\label{fig:issue-def}
\centering
\end{figure}

Semantic segmentation is a fundamental technique for numerous computer vision applications like scene understanding, human parsing and autonomous driving. With the recent development of the convolutional neural network, especially the Fully Convolutional Network (FCN)~\cite{Long-CVPR-FCN-2015}, a lot of great work such as~\cite{Zhao-CVPR-PSPNet-2017, Chen-Arxiv-Deeplabv3-2017, Lin-CVPR-Refinenet-2017,Peng-CVPR-Largekernl-2017} have obtained promising results on the benchmarks. However, the features learned by these methods are usually not discriminative to differentiate 1) the patches which share the same semantic label but different appearances, named intra-class inconsistency as shown in the first row of Figure~\ref{fig:issue-def}; 2) the two adjacent patches which have different semantic labels but with similar appearances, named inter-class indistinction as shown in the second row of Figure~\ref{fig:issue-def}. 

To address these two challenges, we rethink the semantic segmentation task from a more macroscopic point of view. In this way, we regard the semantic segmentation as a task to assign a consistent semantic label to a category of things, rather than to each single pixel. From a macroscopic perspective, regarding each category of pixels as a whole, inherently considers both intra-class consistency and inter-class variation. It means that the task demands discriminative features. To this end, we present a novel Discriminative Feature Network (DFN) to learn the feature representation which considers both the ``intra-class consistency'' and the ``inter-class distinction''.

\begin{figure*}[t]
\label{fig:network}
\includegraphics[width=\linewidth]{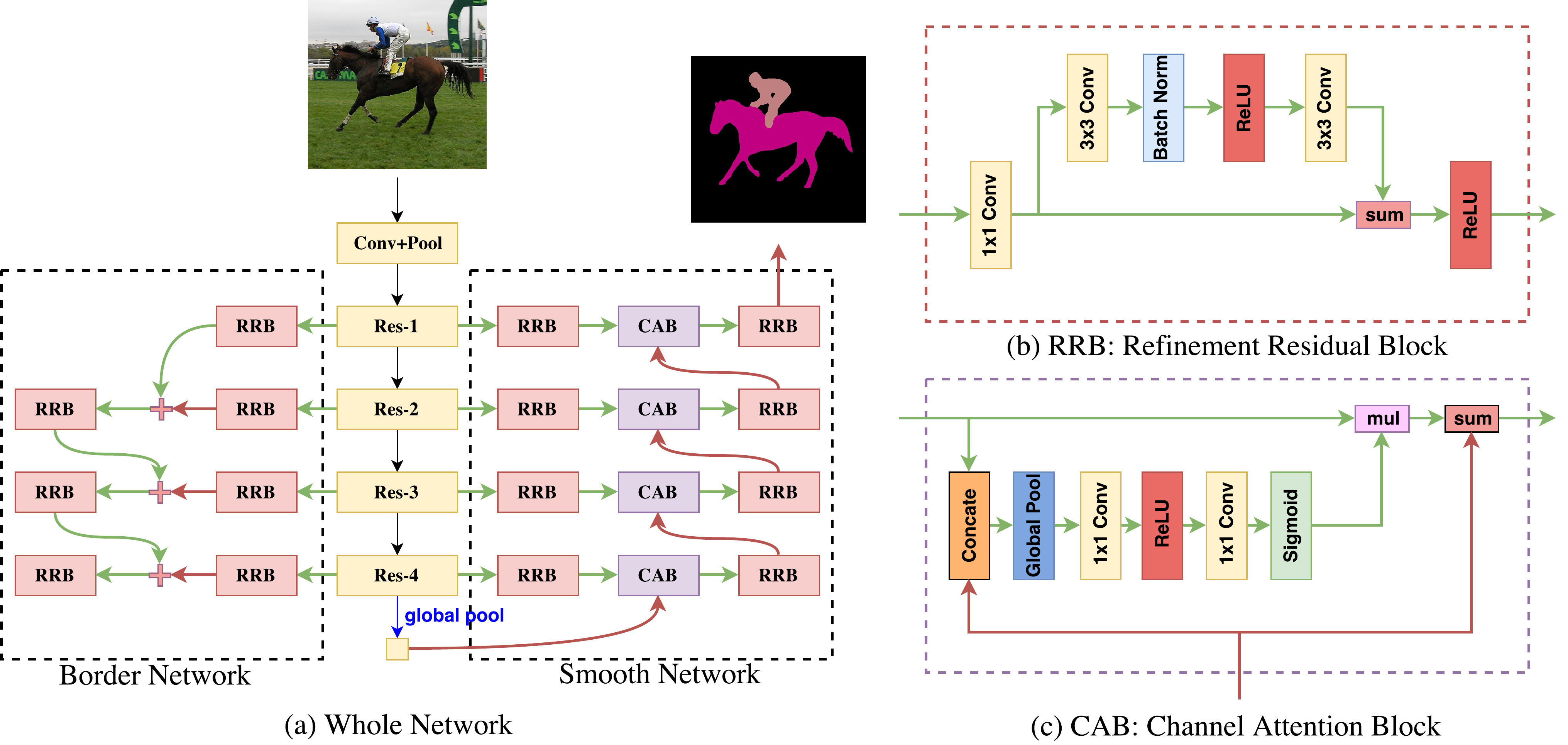}
\vspace{-4mm}
\caption{An overview of the Discriminative Feature Network. 
(a) Network Architecture.
(b) Components of the Refinement Residual Block (RRB).
(c) Components of the Channel Attention Block (CAB).
The red and blue lines represent the upsample and downsample operators, respectively.
The green line can not change the size of feature maps, just a path of information passing.}
\centering
\end{figure*}

Our DFN involves two components: Smooth Network and Border Network, as Figure~\ref{fig:network} illustrates. The Smooth Network is designed to address the intra-class inconsistency issue. To learn a robust feature representation for intra-class consistency, we usually consider two crucial factors. On the one hand, we need multi-scale and global context features to encode the local and global information. For example, the small white patch only in Figure~\ref{fig:issue-def}(a) usually cannot predict the correct category due to the lack of sufficient context information. On the other hand, as multi-scale context is introduced, for a certain scale of thing, the features have different extent of discrimination, some of which may predict a false label. Therefore, it is necessary to select the discriminative and effective features. Motivated by these two aspects, our Smooth Network is presented based on the U-shape~\cite{Peng-CVPR-Largekernl-2017, Lin-CVPR-Refinenet-2017, Ronneberger-ICCV-U-net-2015, Ghiasi-ECCV-LRR-2016, Xie-ICCV-HEAD-2015} structure to capture the multi-scale context information, with the global average pooling~\cite{Lin-ICLR-NIN-2013, Liu-ICLR-ParseNet-2016, Zhao-CVPR-PSPNet-2017, Chen-Arxiv-Deeplabv3-2017} to capture the global context. Also, we propose a Channel Attention Block (CAB), which utilizes the high-level features to guide the selection of low-level features stage-by-stage.  

Border Network, on the other hand, tries to differentiate the adjacent patches with similar appearances but different semantic labels. Most of the existing approaches~\cite{Liu-ICLR-ParseNet-2016, Zhao-CVPR-PSPNet-2017, Chen-Arxiv-Deeplabv3-2017, Peng-CVPR-Largekernl-2017} consider the semantic segmentation task as a dense recognition problem, which usually ignores explicitly modeling the inter-class relationship. Consider the example in Figure~\ref{fig:issue-def}(d), if more and more global context is integrated into the classificiation process, the computer case next to the monitor can be easily misclassified as a monitor due to the similar appearance. Thus, it is significant to explicitly involve the semantic boundary to guide the learning of the features. It can amplify the variation of features on both sides. In our Border Network, we integrate semantic boundary loss during the training process to learn the discriminative features to enlarge the ``inter-class distinction''.

In summary, there are four contributions in our paper:
\begin{compactitem}
    \item We rethink the semantic segmentation task from a new macroscopic point of view. We regard the semantic segmentation as a task to assign a consistent semantic label to one category of things, not just at the pixel level.
	\item We propose a Discriminative Feature Network to simultaneously address the ``intra-class consistency'' and ``inter-class variation'' issues. Experiments on PASCAL VOC 2012 and Cityscapes datasets validate the effectiveness of our proposed algorithm.
	\item We present a Smooth Network to enhance the intra-class consistency with the global context and the Channel Attention Block.
	\item We design a bottom-up Border Network with deep supervision to enlarge the variation of features on both sides of the semantic boundary. This can also refine the semantic boundary of prediction.
\end{compactitem}

\section{Related Work}
\label{sec:related_work}
Recently, lots of approaches based on FCN have achieved high performance on different benchmarks~\cite{Zhou-ADE-2017,Pascal-VOC-2012,Cityscapes}.  Most of them are still constrained by intra-class inconsistency and inter-class indistinction issues. 

\vspace{-2ex}    
\paragraph{Encoder-Decoder:} The FCN model has inherently encoded different levels of feature. Naturally, some methods integrate them to refine the final prediction. This branch of methods mainly consider how to recover the reduced spatial information caused by consecutive pooling operator or convolution with stride. For example, SegNet~\cite{Badrinarayanan-PAMI-SegNet-2017} utilizes the saved pool indices to recover the reduced spatial information. U-net~\cite{Ronneberger-ICCV-U-net-2015} uses the skip connection, while the Global Convolutional Network~\cite{Peng-CVPR-Largekernl-2017} adapts the large kernel size. Besides, LRR~\cite{Ghiasi-ECCV-LRR-2016} adds the Laplacian Pyramid Reconstruction network, while RefineNet~\cite{Lin-CVPR-Refinenet-2017} utilizes multi-path refinement network. However, this type of architecture ignores the global context. In addition, most methods of this type are just summed up the features of adjacent stages without consideration of their diverse representation. This leads to some inconsistent results.  

\vspace{-2ex}    
\paragraph{Global Context:} Some modern methods have proven the effectiveness of global average pooling. ParseNet~\cite{Liu-ICLR-ParseNet-2016} firstly applies global average pooling in the semantic segmentation task. Then PSPNet~\cite{Zhao-CVPR-PSPNet-2017} and Deeplab v3~\cite{Chen-Arxiv-Deeplabv3-2017} respectively extend it to the Spatial Pyramid Pooling~\cite{He-ECCV-SPP-2014} and Atrous Spatial Pyramid Pooling~\cite{Chen-Arxiv-Deeplabv2-2016}, resulting in great performance in different benchmarks. However, to take advantage of the pyramid pooling module sufficiently, these two methods adopt the base feature network to 8 times down-sample with atrous convolution~\cite{Chen-Arxiv-Deeplabv2-2016, Yu-ICLR-Dilate-2016}, which is time-consuming and memory intensive.
  
\vspace{-2ex}    
\paragraph{Attention Module:} Attention is helpful to focus on what we want. Recently, the attention module becomes increasingly a powerful tool for deep neural networks~\cite{Mnih-NIPS-RecurrentAttention-2014, Wang-CVPR-ResAttention-2017, Hu-Arxiv-SEnet-2017, Chen-CVPR-SCACNN-2017}. The method in \cite{Chen-CVPR-AttentionScale-2016} pays attention to different scale information. In this work, we utilize channel attention to select the features similar to SENet~\cite{Hu-Arxiv-SEnet-2017}.

\vspace{-2ex}    
\paragraph{Semantic Boundary Detection:} Boundary detection is a fundamental challenge in computer vision. There are lots of specific methods proposed for the task of boundary detection~\cite{Yu-CVPR-CASENet-2017, Xie-ICCV-HEAD-2015, Yang-CVPR-Contour-2016, Liu-CVPR-RCF-2017}. Most of these methods straightly concatenate the different level of features to extract the boundary. However, in this work, our goal is to obtain the features with inter-class distinction as much as possible with accurate boundary supervision. Therefore, we design a bottom-up structure to optimize the features on each stage.


\section{Method}
\label{sec:method}
In this section, we first detailedly introduce our proposed Discriminative Feature Network containing Smooth Network and Border Network. Then, we elaborate how these two networks specifically handle the intra-class consistency issue and the inter-class distinction issue. Finally, we describe the complete encoder-decoder network architecture, Discriminative Feature Network. 

\subsection{Smooth network}
\label{sec:sn}
In the task of semantic segmentation, most of modern methods consider it as a dense prediction issue. However, the prediction sometimes has incorrect results in some parts, especially the parts of large regions and complex scenes, which is named intra-class inconsistency issue. 

The intra-class inconsistency problem is mainly due to the lack of context. Therefore, we introduce the global context with global average pooling~\cite{Liu-ICLR-ParseNet-2016, Lin-ICLR-NIN-2013, Zhao-CVPR-PSPNet-2017, Chen-Arxiv-Deeplabv3-2017}. However, global context just has the high semantic information, which is not helpful for recovering the spatial information. Consequently, we further need the multi-scale receptive view and context to refine the spatial information, as most modern approaches~\cite{Zhao-CVPR-PSPNet-2017,Chen-Arxiv-Deeplabv3-2017, Peng-CVPR-Largekernl-2017} do. Nevertheless, there exists a problem that the different scales of receptive views produce the features with different extents of discrimination, leading to inconsistent results. Therefore, we need to select more discriminative features to predict the unified semantic label of one certain category. 

In our proposed network, we use ResNet~\cite{He-CVPR-ResNet-2016} as a base recognition model. This model can be divided into five stages according to the size of the feature maps. According to our observation, the different stages have different recognition abilities resulting in diverse consistency manifestation. In the lower stage, the network encodes finer spatial information, however, it has poor semantic consistency because of its small receptive view and without the guidance of spatial context. While in the high stage, it has strong semantic consistency due to large receptive view, however, the prediction is spatially coarse. Overall, the lower stage makes more accurate spatial predictions, while the higher stage gives more accurate semantic predictions. Based on this observation, to combine their advantages, we propose a Smooth Network to utilize the high stage's consistency to guide the low stage for the optimal prediction.

\begin{figure}[t]
\subfigure[Channel Attention Block]{
\label{fig:cab}
\includegraphics[width=\linewidth]{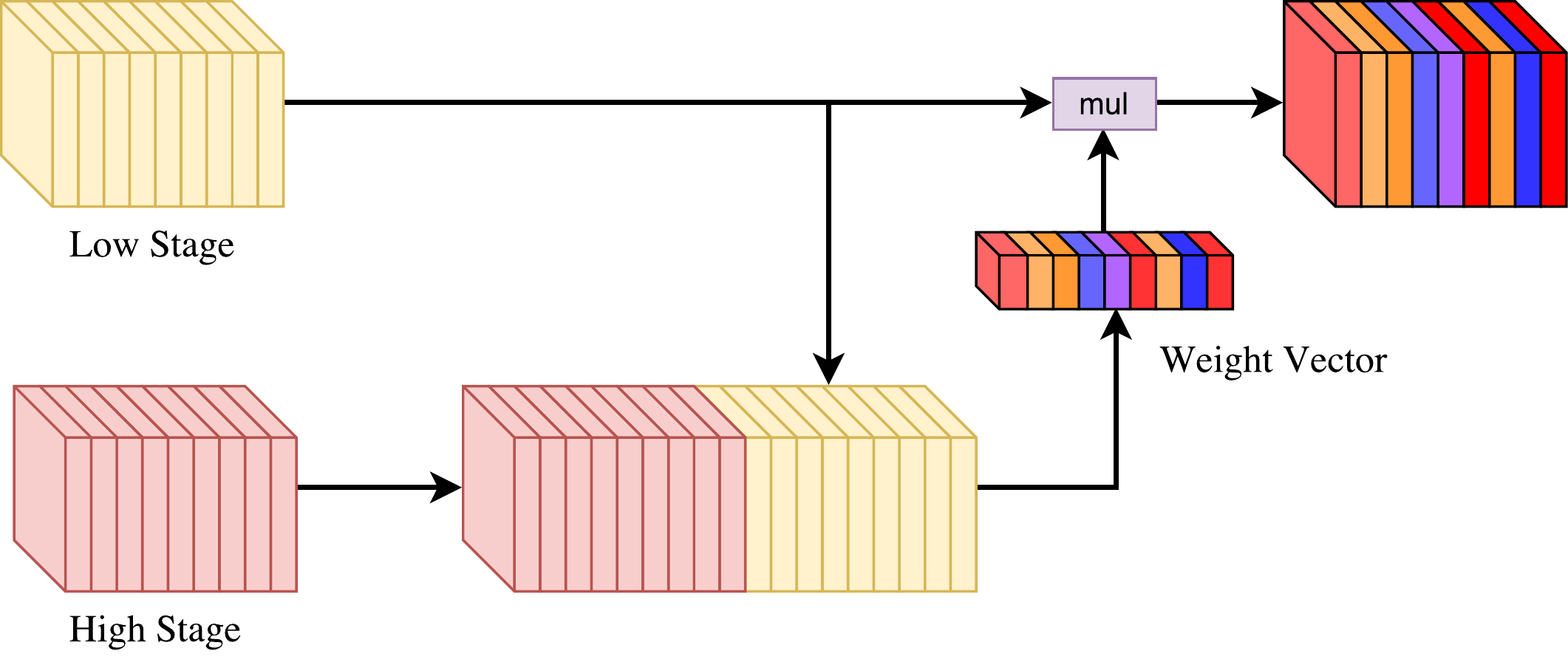}
}
\subfigure[Attention Vector]{
\label{fig:av}
\includegraphics[width=\linewidth]{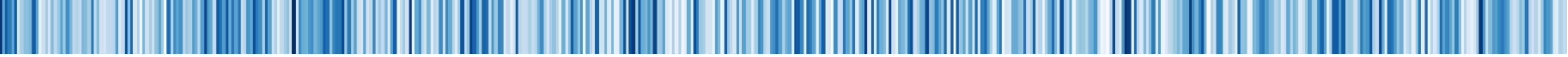}
}
\caption{Schematic diagram of Channel Attention Block. In (a), the yellow block represents the feature of low stage, while the red one represents high stage. We concatenate the features of adjacent stages to compute a weight vector, which re-weights the feature maps of low stage. The hotter color represents the high weight value. In (b), it is the real attention value vector from the stage-4 channel attention block. The deeper blue represents the higher weight value.}
\label{fig:channel-attetion}
\centering
\end{figure}

We observe that in the current prevalent semantic segmentation architecture, there are mainly two styles. The first one is ``Backbone-Style'', such as PSPNet~\cite{Zhao-CVPR-PSPNet-2017}, Deeplab v3~\cite{Chen-Arxiv-Deeplabv3-2017}. It embeds different scale context information to improve the consistency of network with the Pyramid Spatial Pooling module~\cite{He-ECCV-SPP-2014} or Atrous Spatial Pyramid Pooling module~\cite{Chen-Arxiv-Deeplabv2-2016}. The other one is ``Encoder-Decoder-Style'', like RefineNet~\cite{Lin-CVPR-Refinenet-2017}, Global Convolutional Network~\cite{Peng-CVPR-Largekernl-2017}. This style of network utilizes the inherent multi-scale context of different stage, but it lacks the global context which has the strongest consistency. In addition, when the network combines the features of adjacent stages, it just sums up these features by channel. This operation ignores the diverse consistency in different stages. To remedy the defect, we first embed a global average pooling layer~\cite{Liu-ICLR-ParseNet-2016} to extend the U-shape architecture~\cite{Long-CVPR-FCN-2015, Xie-ICCV-HEAD-2015} to a V-shape architecture. With the global average pooling layer, we introduce the strongest consistency constraint into the network as a guidance. Furthermore, to enhance consistency, we design a Channel Attention Block, as shown in Figure~\ref{fig:network}(c). This design combines the features of adjacent stages to compute a channel attention vector~\ref{fig:av}. The features of high stage provide a strong consistency guidance, while the features of low stage give the different discrimination information of features. In this way, the channel attention vector can select the discriminative features. 

\vspace{-2ex}    
\paragraph{Channel attention block:} Our Channel Attention Block (CAB) is designed to change the weights of the features on each stage to enhance the consistency, as illustrated in Figure~\ref{fig:channel-attetion}. In the FCN architecture, the convolution operator outputs a score map, which gives the probability of each class at each pixel. In Equation~\ref{eqn:eqn-sum}, the final score at score map is just summed over all channels of feature maps.

\begin{equation}
\label{eqn:eqn-sum}
y_k = F(x; w) = \sum_{i=1,j=1}^{D}w_{i,j}x_{i,j}
\end{equation}
where $x$ is the output feature of network. $w$ represents the convolution kernel. And $k \in \{1, 2, \ldots, K\}$. $K$ is the number of channels. $D$ is the set of pixel positions.

\begin{equation}
	\label{eqn:pred}
	\delta_{i}(y_k) = \frac{exp(y_k)}{\sum_{j=1}^K{exp(y_j)}}
\end{equation}
where $\delta$ is the prediction probability. $y$ is the output of network.

As shown in Equation~\ref{eqn:eqn-sum} and Equation~\ref{eqn:pred}, the final predicted label is the category with highest probability. Therefore, we assume that the prediction result is $y_0$ of a certain patch, while its true label is $y_1$. Consequently, we can introduce a parameter $\alpha$ to change the highest probability value from $y_0$ to $y_1$, as Equation~\ref{eqn:eqn-re-weight} shows.  
\begin{equation}
\label{eqn:eqn-re-weight}
\bar{y} = \alpha y = \begin{bmatrix}\alpha_1 \\ \vdots \\ \alpha_K \end{bmatrix} \cdot \begin{bmatrix}y_1 \\ \vdots \\ y_K \end{bmatrix}=\begin{bmatrix}\alpha_1 w_1 \\ \vdots \\ \alpha_K w_K \end{bmatrix} \times \begin{bmatrix}x_1 \\ \vdots \\ x_K \end{bmatrix}
\end{equation}
where $\bar{y}$ is the new prediction of network and $\alpha = Sigmoid(x;w)$

Based on the above formulation of the Channel Attention Block (CAB), we can explore its practical significance. In Equation~\ref{eqn:eqn-sum}, it implicitly indicates that the weights of different channels are equal. However, as mentioned in Section~\ref{sec:introduction}, the features in different stages have different degrees of discrimination, which results in different consistency of prediction. In order to obtain the intra-class consistent prediction, we should extract the discriminative features and inhibit the indiscriminative features. Therefore, in Equation~\ref{eqn:eqn-re-weight}, the $\alpha$ value applies on the feature maps $x$, which represents the feature selection with CAB. With this design, we can make the network to obtain discriminative features stage-wise  to make the prediction intra-class consistent.

\vspace{-2ex}    
\paragraph{Refinement residual block:} The feature maps of each stage in feature network all go through the Refinement Residual Block, schematically depicted in Figure~\ref{fig:network}(b). The first component of the block is a $1 \times 1$ convolution layer. We use it to unify the number of channels to 512. Meanwhile, it can combine the information across all channels. Then the following is a basic residual block, which can refine the feature map. Furthermore, this block can strengthen the recognition ability of each stage, inspired from the architecture of ResNet~\cite{He-CVPR-ResNet-2016, He-ECCV-Identity-2016}. 
	
\subsection{Border network}
In the semantic segmentation task, the prediction is confused with the different categories with similar appearances, especially when they are adjacent spatially. Therefore, we need to amplify the distinction of features. With this motivation, we adopt a semantic boundary to guide the feature learning. To extract the accurate semantic boundary, we apply the explicit supervision of semantic boundary, which makes the network learn a feature with strong inter-class distinctive ability.  Therefore, we propose a Border Network to enlarge the inter-class distinction of features. It directly learns a semantic boundary with an explicit semantic boundary supervision, similar to a semantic boundary detection task. This makes the features on both sides of semantic boundary distinguishable.

As stated in Section~\ref{sec:sn}, the feature network has different stages. The low stage features have more detailed information, while the high stage features have higher semantic information. In our work, we need semantic boundary with more semantic meanings. Therefore, we design a bottom-up Border Network. This network can simultaneously get accurate edge information from low stage and obtain semantic information from high stage, which eliminates some original edges lack of semantic information. In this way, the semantic information of high stage can refine the detailed edge information from low stage stage-wise. The supervisory signal of the network is obtained from the semantic segmentation's groundtruth with a traditional image processing method, such as Canny~\cite{Canny-PAMI-Canny-1986}. 

To remedy the imbalance of the positive and negative samples, we use focal loss~\cite{Lin-ICCV-Focal-2017} to supervise the output of the Border Network, as shown in Equation~\ref{eqn:eqn-focal-loss}. We adjust the parameters $\alpha$ and $\gamma$ of focal loss for better performance.
\begin{equation}
\label{eqn:eqn-focal-loss}
FL(p_k) = -(1-p_k)^{\gamma}\log{p_k}
\end{equation}
where $p_k$ is the estimated probability for class $k$, $k \in \{1, 2, \ldots, K\}$. And $K$ is the maximum value of class label.

The Border Network mainly focuses on the semantic boundary which separates the classes on two sides of the boundary. For extracting accurate semantic boundary, the features on both sides will become more distinguishable. This exactly reaches our goal to make the features with inter-class distinction as much as possible.

\subsection{Network Architecture}
With Smooth Network and Border Network, we propose our Discriminative Feature Network for semantic segmentation as illustrated in Figure~\ref{fig:network}~(a).

We use pre-trained ResNet~\cite{He-CVPR-ResNet-2016} as a base network. In the Smooth Network, we add the global average pooling layer on the top of the network to get the strongest consistency. Then we utilize the channel attention block to change the weights of channels to further enhance the consistency. Meanwhile, in the Border Network, with the explicit semantic boundary supervision, the network obtains accurate semantic boundary and makes the bilateral features more distinct. With the support of both sub-networks, the intra-class features become more consistent, while the inter-class ones grow more distinct. 

For explicit feature refinement, we use deep supervision to get better performance and make the network easier to optimize. In the Smooth Network, we use the softmax loss to supervise the each stage's upsampled output excluding the global average pooling layer, while we use the focal loss to supervise the outputs of Border Network. Finally, we use a parameter $\lambda$ to balance the segmentation loss $\ell_{s}$ and the boundary loss $\ell_{b}$, as Equation~\ref{eqn:eqn-loss} shows.

\begin{align}
\ell_{s} &= SoftmaxLoss(y;w) \\
\ell_{b} &= FocalLoss(y;w) \\
L &= \ell_{s}+\lambda \ \ell_{b}
\label{eqn:eqn-loss}
\end{align}

\section{Experimental Results}
\label{sec:experiments}
We evaluate our approach on two public datasets: PASCAL VOC 2012~\cite{Pascal-VOC-2012} and Cityscapes~\cite{Cityscapes}. We first introduce the datasets and report the implementation details. Then we evaluate each component of the proposed method, and analyze the results in detail. Finally, we present the comparison results with other state-of-the-art methods.

\vspace{-2ex}    
\paragraph{PASCAL VOC 2012:} The PASCAL VOC 2012 is a well-known semantic segmentation benchmark which contains 20 object classes and one background, involving 1,464 images for training, 14,449 images for validation and 1,456 images for testing. The original dataset is augmented by the Semantic Boundaries Dataset~\cite{Hariharan-SBD-2011}, resulting in 10,582 images for training. 

\vspace{-2ex}
\paragraph{Cityscapes:} The Cityscapes is a large semantic segmentation dataset of urban street scene in car perspective. The dataset contains 30 classes, of which 19 classes are considered for training and evaluation. There are 2,979 images for training, 500 images for validation and 1,525 images for testing, which are all fine annotated. And there are another 19,998 images with coarse annotation. The images all have a high resolution of 2,048$\times$1,024.

\subsection{Implementation details}

Our proposed network is based on the ResNet-101 pre-trained on Image{N}et~\cite{ILSVRC15}. And we use the FCN4~\cite{Long-CVPR-FCN-2015, Xie-ICCV-HEAD-2015} as our base segmentation framework.

\vspace{-2ex}    
\paragraph{Training:} We train the network using mini-batch stochastic gradient descent (SGD)~\cite{Krizhevsky-NIPS-Imagenet} with batch size $32$, momentum $0.9$ and weight decay $0.0001$. Inspired by \cite{Chen-Arxiv-Deeplabv2-2016, Liu-ICLR-ParseNet-2016}, we use the ``poly'' learning rate policy where the learning rate is multiplied by $\left(1- \frac{iter}{max\_iter}\right)^{power}$ with power $0.9$ and initial learning rate $4e^{-3}$. As for the $\lambda$, we finally use the value of $0.1$ after a series of comparison experiments. For measuring the performance of our proposed network, we use the mean pixel intersection-over-union (mean IOU) as the metric. 

\vspace{-2ex}    
\paragraph{Data augmentation:} We use mean subtraction and random horizontal flip in training for both PASCAL VOC 2012 and Cityscapes. In addition, we find it is crucial to randomly scale the input images, which improves the performance obviously. We use 5 scales $\{0.5, 0.75, 1, 1.5, 1.75\}$ on both datasets.

\subsection{Ablation study}
In this subsection, we will step-wise decompose our approach to reveal the effect of each component. In the following experiments, we evaluate all comparisons on PASCAL VOC 2012 dataset~\cite{Pascal-VOC-2012}. And we report the comparison results in PASCAL VOC 2012 dataset~\cite{Pascal-VOC-2012} and Cityscapes dataset~\cite{Cityscapes}.

\begin{figure}[t]
\centering
\includegraphics[width=\linewidth]{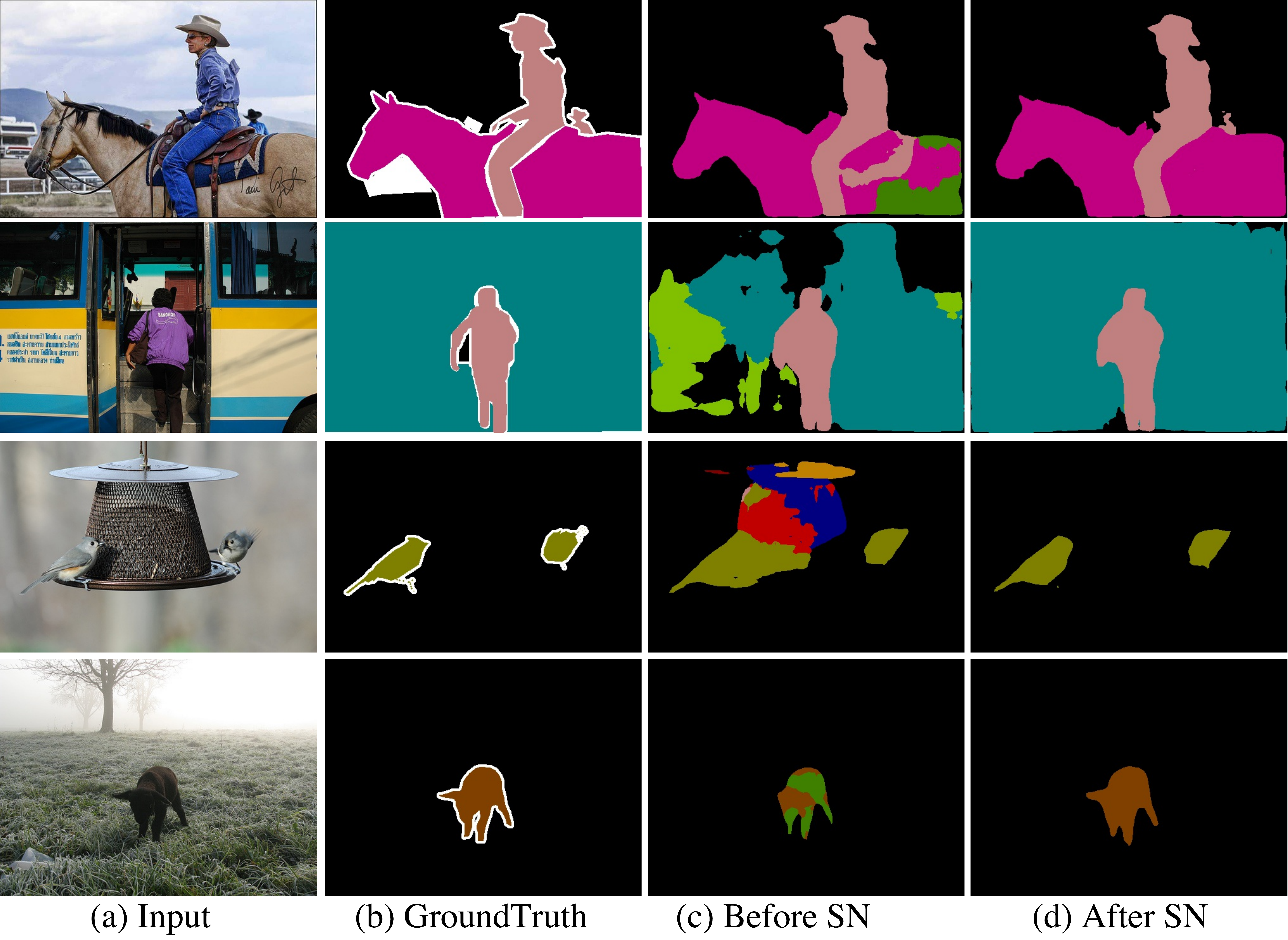}
\caption{Results of Smooth Network on PASCAL VOC 2012 dataset. }
\label{fig:result-sn}
\end{figure}

\begin{table}[t]
\begin{center}
\caption{The performance of ResNet-101 with and without random scale augmentation.}
\label{tab:baseline}
\begin{tabular}{lcc}
\toprule
Method & Random\_Scale & Mean IOU(\%)\\
\hline
 \noalign{\smallskip}
Res-101 & & 69.26 \\
Res-101 & $\surd$ & 72.86\\
\bottomrule
\end{tabular}
\end{center}
\end{table}

\subsubsection{Smooth network}
We use the ResNet-101 as our base feature network, and directly upsample the ouput. First, we evaluate the performance of the base ResNet-101, as shown in Table~\ref{tab:baseline}. Then we extend the base network to FCN4 structure~\cite{Long-CVPR-FCN-2015, Xie-ICCV-HEAD-2015} with our proposed Refinement Residual Block (RRB), which improves the performance from $72.86\%$ to $76.65\%$, as Table~\ref{tab:combine} shows. We visualize the effect of the Smooth Network. Figure~\ref{fig:result-sn} presents some examples of semantic segmentation results. Obviously, our Smooth Network can effectively make the prediction more consistent.

\vspace{-2ex}    
\paragraph{Ablation for global pooling:} We need the features with strong consistency. Thus based our observation in Section~\ref{sec:method}, we add the global average pooling on the top of the network. As shown in Table~\ref{tab:combine}, the global average pooling introduces the strongest consistency to guide other stages. This improves the performance from $76.65\%$ to $78.20\%$, which is an obvious improvement.

\begin{figure}[t]
\centering
\includegraphics[width=\linewidth]{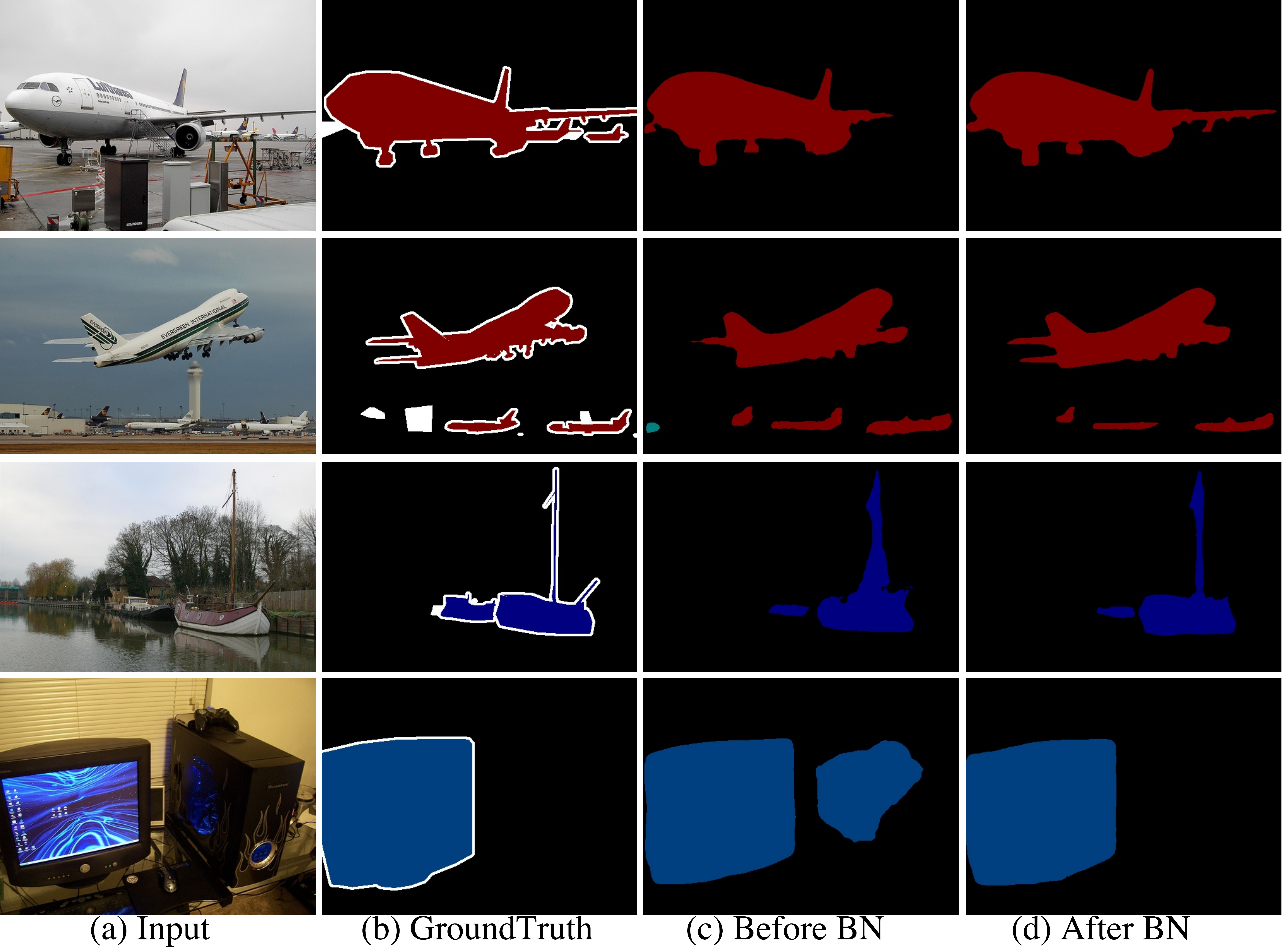}
\caption{Results of Border Network on PASCAL VOC 2012 dataset. The boundary on prediction is refined by the Border Network.}
\label{fig:result-bn}
\end{figure}

\begin{table}[t]
\begin{center}
\caption{Detailed performance comparison of our proposed Smooth Network. \textbf{RRB:} refinement residual block. \textbf{GP:} global pooling branch. \textbf{CAB:} channel attention block. \textbf{DS:} deep supervision. }
\label{tab:combine}
\begin{tabular}{lc}
\toprule
Method & Mean IOU(\%)\\
\hline
 \noalign{\smallskip}
Res-101 & 72.86\\
Res-101+RRB & 76.65\\
\hline
 \noalign{\smallskip}
Res-101+RRB+GP & 78.20\\
Res-101+RRB+GP+CAB & 79.31\\
\hline
 \noalign{\smallskip}
Res-101+RRB+DS & 77.08\\
Res-101+RRB+GP+DS & 78.51\\
Res-101+RRB+GP+CAB+DS & 79.54\\
\bottomrule
\end{tabular}
\end{center}
\end{table}

\vspace{-2ex}    
\paragraph{Ablation for deep supervision:} To refine the hierarchical features, we use deep supervision. We add the softmax loss on each stage excluding the global average pooling layer. As shown in Table~\ref{tab:combine}, this further improves the performance by almost $0.4\%$.

\vspace{-2ex}    
\paragraph{Ablation for channel attention block:} Based on the aforementioned architecture, we add the Channel Attention Block~(CAB). It utilizes the high stage to guide the low stage with a channel attention vector to enhance consistency, which improves the performance from $78.51\%$ to $79.54\%$ over evaluation, as Table~\ref{tab:combine} shows.

\begin{figure}[t]
\centering
\includegraphics[width=\linewidth]{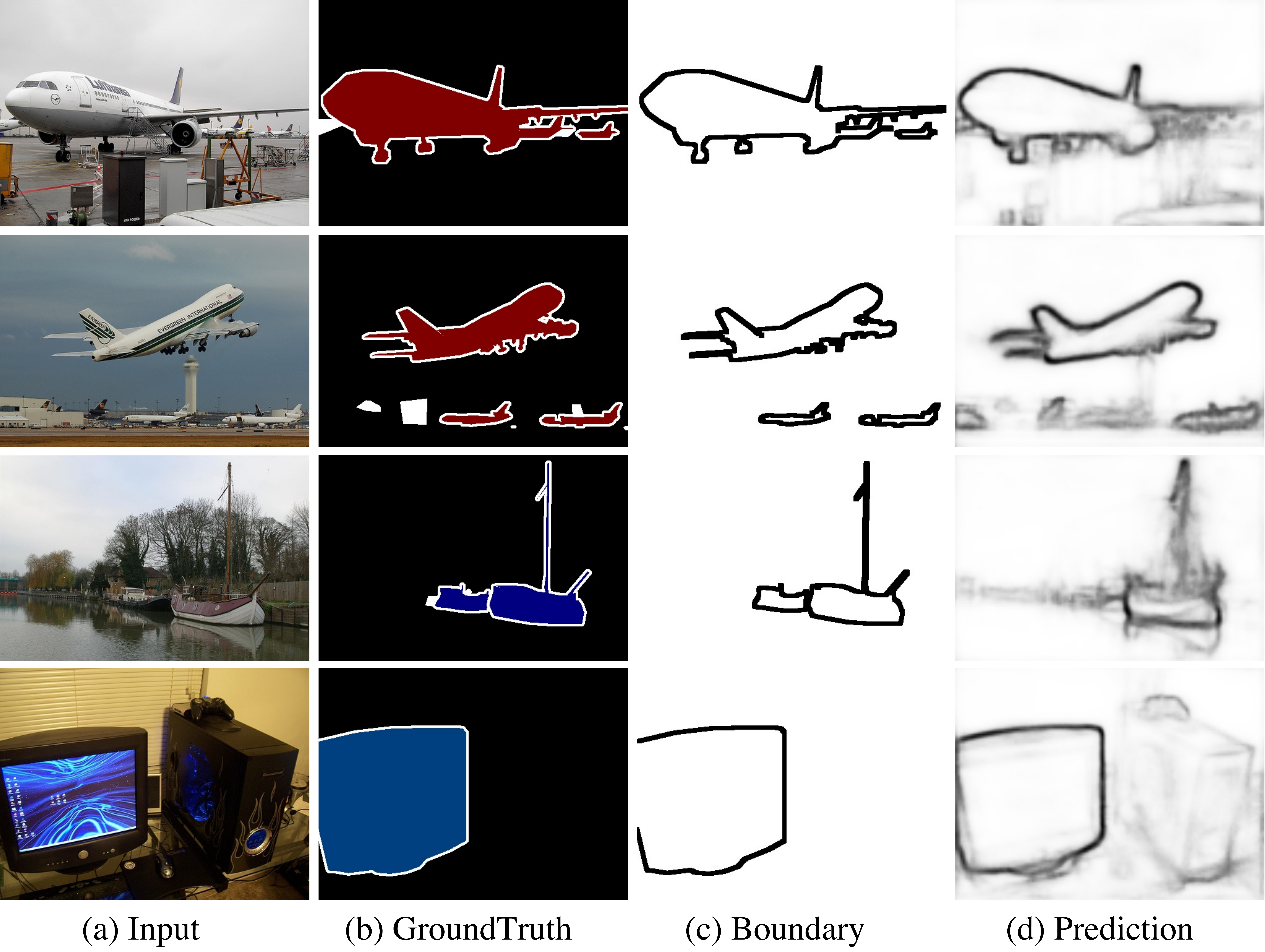}
\caption{The boundary prediction of Border Network on PASCAL VOC 2012 dataset. The third column is the semantic boundary extracted from GroundTruth by Canny operator. The last column is the prediction results of Border Network.}
\label{fig:result-bn-edge}
\end{figure}

\subsubsection{Border network}
While the Smooth Network pays attention to the intra-class consistency, the Border Network focuses on the inter-class indistinction. Due to the accurate boundary supervisory signal, the network amplifies the distinction of bilateral feature to extract the semantic boundary. Then we integrate the Border Network into the Smooth Network. This improves the performance from $79.54\%$ to $79.67\%$, as shown in Table~\ref{tab:bn}. The Border Network optimizes the semantic boundary, which is a comparably small part of the whole image, so this design makes a minor improvement. We visualize the effect of Border Network, as shown in Figure~\ref{fig:result-bn}. In addition, Figure~\ref{fig:result-bn-edge} shows the predicted semantic boundary of Border Network. We can obviously observe that the Border Network can focus on the semantic boundary preferably. 

\begin{table}[t]
\begin{center}
\caption{Combining the Border Network and Smooth Network as Discriminative Feature Network. \textbf{SN}: Smooth Network. \textbf{BN:} Border Network. \textbf{MS\_Flip:} Adding multi-scale inputs and left-right flipped inputs.}
\label{tab:bn}
\begin{tabular}{lc}
\toprule
Method & Mean IOU(\%)\\
\hline
\noalign{\smallskip}
Res-101+SN & 79.54\\
Res-101+SN+BN & 79.67\\
\hline
\noalign{\smallskip}
Res-101+SN+MS\_Flip & 79.90\\
Res-101+SN+BN+MS\_Flip & 80.01\\
\bottomrule
\end{tabular}
\end{center}
\end{table}

\subsubsection{Discriminative Feature network}
With the Discriminative Feature Network (DFN), we conduct experiments about the balance parameter of the combined loss. Then we present the final results on PASCAL VOC 2012 and Cityscapes datasets.

\begin{figure}[t]
	\centering
	\includegraphics[width=\linewidth]{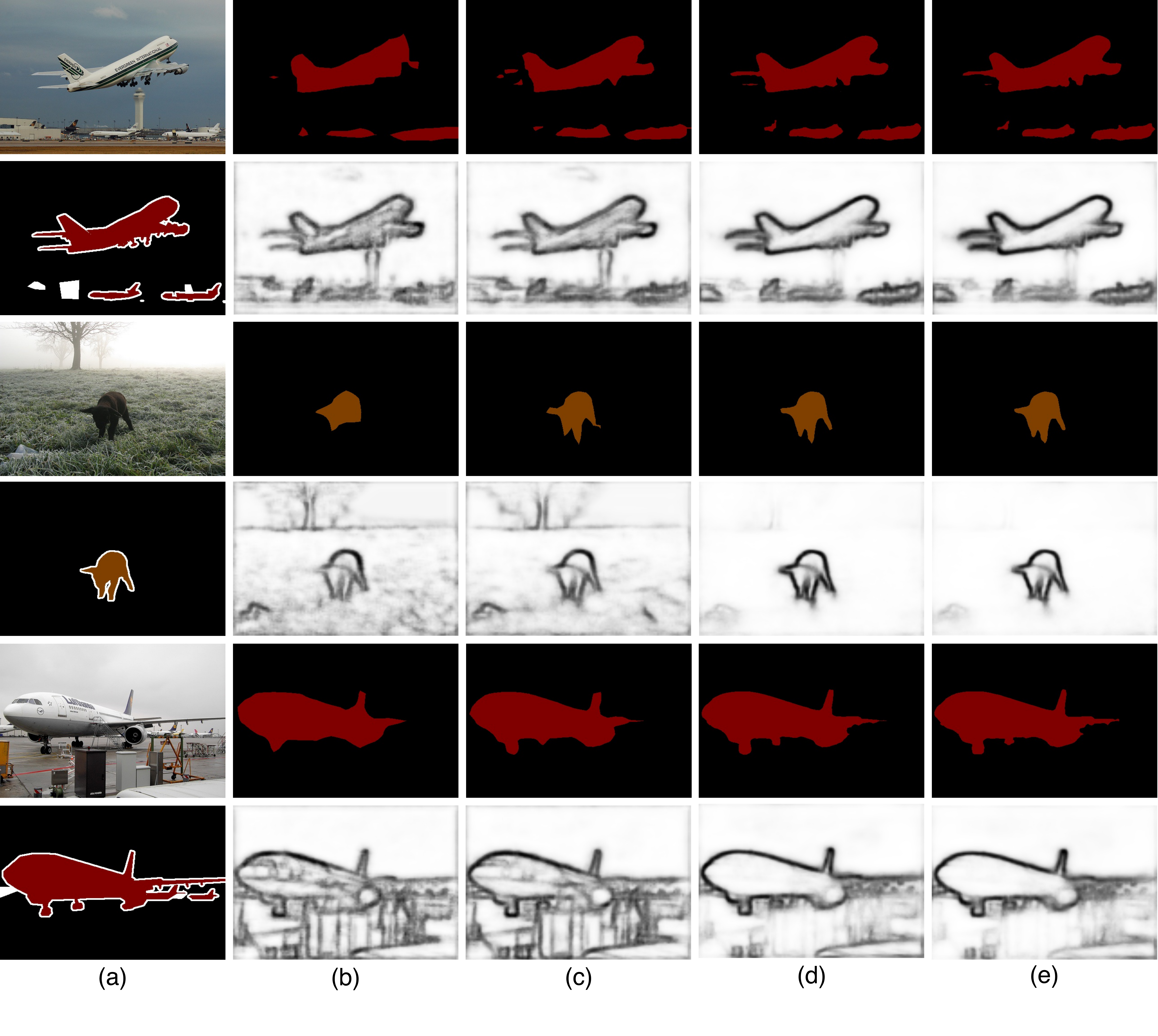}
	\caption{Example results of DFN in the stage-wise refinement process on PASCAL VOC 2012 dataset. The first column is the original image and groundtruth. The last is the refinement process of two networks. The segmentation prediction in lower stage is more spatial coarse, and the higher is finer.  While the boundary prediction in lower stage contains more edges not belong to semantic boundary, the semantic boundary in higher stage is more pure.}
	\label{fig:refine}
\end{figure}

\begin{figure}[t]
\centering
\includegraphics[width=\linewidth]{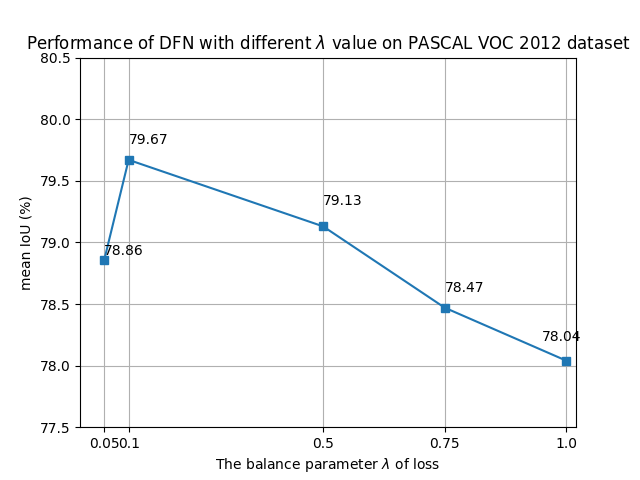}
\caption{Results of DFN with different $\lambda$ value on PASCAL VOC 2012 dataset.}
\label{fig:lambda}
\end{figure}

\vspace{-2ex}    
\paragraph{Balance of both losses:} The balance weight between the losses of two networks is crucial. To further analyze the effect of these two networks, we conduct experiments for different balance value. We test five values of $\{0.05, 0.1, 0.5, 0.75, 1\}$. As shown in Figure~\ref{fig:lambda}, with the same setting, our method achieves the highest performance with the value of $0.1$. 

\vspace{-2ex}    
\paragraph{Stage-wise refinement:}
It is worth noting that both Smooth Network and Border Network use the stage-wise mechanism. The Smooth Network utilizes a top-down stage-wise manner to transmit the context information from high stage to low stage, to ensure the inter-class consistency. On the other hand, the Border Network uses a bottom-up stage-wise manner to refine the semantic boundary with the edge information in the lower stage. With the bidirectional stage-wise mechanism, the Smooth Network and Border Network respectively refine the segmentation and boundary prediction, as shown in Figure~\ref{fig:refine}. The gradually accurate predictions validate the effectiveness of the stage-wise mechanism. 

\begin{table}
\begin{center}
\caption{Validation strategy on PASCAL VOC 2012 dataset. \textbf{MS\_Flip:} Multi-scale and flip evaluation.}
\label{tab:evalres}
\begin{tabular}{cccc}
\toprule
Method & train\_data & MS\_Flip & Mean IOU(\%)\\
\hline
\noalign{\smallskip}
DFN & & & 79.67\\
DFN & $\surd$ & & 80.46\\
DFN & $\surd$ & $\surd$ & 80.60\\
\bottomrule
\end{tabular}
\end{center}
\end{table}

 \begin{table}[t]
 \begin{center}
 \caption{Performance on PASCAL VOC 2012 test set. Methods pre-trained on MS-COCO are marked with $^+$.}
 \label{tab:vocresult}
 \begin{tabular}{lc}
 \toprule
 Method & Mean IOU(\%)\\
 \hline
   \noalign{\smallskip}
 FCN~\cite{Long-CVPR-FCN-2015} & 62.2\\
 Zoom-out~\cite{Mostajabi-CVPR-Zoom-2015} & 69.6 \\
 ParseNet~\cite{Liu-ICLR-ParseNet-2016} & 69.8\\
 Deeplab v2-CRF~\cite{Chen-Arxiv-Deeplabv2-2016} & 71.6 \\
 DPN~\cite{Liu-ICCV-DPN-2015} & 74.1 \\
 Piecewise~\cite{Lin-CVPR-Piecewisetraining-2016} & 75.3\\
 LRR-CRF~\cite{Ghiasi-ECCV-LRR-2016} & 75.9\\
 PSPNet~\cite{Zhao-CVPR-PSPNet-2017} & 82.6\\
 \hline
   \noalign{\smallskip}
 Ours & \textbf{82.7}\\ 
 \hline
 \hline
   \noalign{\smallskip}
 DLC$^+$~\cite{Li-CVPR-DLC-2017} & 82.7 \\
 DUC$^+$~\cite{Wang-CVPR-DUC-2017} & 83.1 \\
 GCN$^+$~\cite{Peng-CVPR-Largekernl-2017} & 83.6 \\
 RefineNet$^+$~\cite{Lin-CVPR-Refinenet-2017} & 84.2 \\
 ResNet-38$^+$~\cite{Wu-Arxiv-Resnet38-2016} & 84.9 \\
 PSPNet$^+$~\cite{Zhao-CVPR-PSPNet-2017} & 85.4 \\
 Deeplab v3$^+$~\cite{Chen-Arxiv-Deeplabv3-2017} & 85.7 \\
 \hline
  \noalign{\smallskip}
 Ours$^+$ & \textbf{86.2} \\
 \bottomrule
 \end{tabular}
 \end{center}
 \end{table}

\vspace{-2ex}    
\paragraph{Performance evaluation on PASCAL VOC 2012:} In evaluation, we apply the multi-scale inputs (with scales $\{0.5, 0.75, 1.0, 1.5, 1.75\}$) and also horizontally flip the inputs to further improve the performance. In addition, since the PASCAL VOC 2012 dataset provides higher quality of annotation than the augmented datasets~\cite{Hariharan-SBD-2011}, we further fine-tune our model on PASCAL VOC 2012 \textit{train} set for evaluation on validation set. More performance details are listed in Table~\ref{tab:evalres}. And then for evaluation on test set, we use the PASCAL VOC 2012 \textit{trainval} set to further fine-tune our proposed method. In the end, our proposed approach respectively achieves performance of $82.7\%$ and $86.2\%$ with and without MS-COCO~\cite{Lin-COCO-2014} fine-tuning, as shown in Table~\ref{tab:vocresult}. Note that, we do not use Dense-CRF~\cite{Chen-ICLR-Deeplab-2016} post-processing for our method.

\begin{figure}[t]
	\centering
	\includegraphics[width=\linewidth]{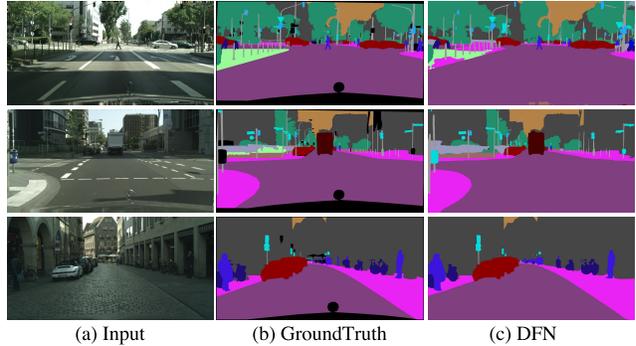}
	\caption{Example results of DFN on Cityscapes dataset.}
	\label{fig:res-city}
\end{figure}

\vspace{-2ex}    
\paragraph{Performance evaluation on Cityscapes:} We also evaluate our approach on the Cityscapes dataset~\cite{Cityscapes}. In training, our crop size of image is $800\times800$. We observe that for the high resolution of image the large crop size is useful. The test performance results are specifically reported in Table~\ref{tab:cityresult}. We visualize the results of our approach on the Cityscapes dataset, as shown in Figure~\ref{fig:res-city}.

 \begin{table}[t]
 \begin{center}
 \caption{Performance on Cityscapes test set. The ``-'' indicates that the method do not present this result in its paper.}
 \label{tab:cityresult}
 \begin{tabular}{lccc}
 \toprule
 \multicolumn{2}{l}{\multirow{2}*{Method}} & \multicolumn{2}{c}{Mean IOU(\%)}\\
 \multicolumn{2}{l}{} & w/o coarse & w/ coarse \\
 \hline
   \noalign{\smallskip}
\multicolumn{2}{l}{CRF-RNN~\cite{Zheng-ICCV-CRFasRNN-2015}} & 62.5 & -\\
 \multicolumn{2}{l}{FCN~\cite{Long-CVPR-FCN-2015}} & 65.3 & -\\
 \multicolumn{2}{l}{DPN~\cite{Liu-ICCV-DPN-2015}} & 66.8 & 59.1 \\
 \multicolumn{2}{l}{LRR~\cite{Ghiasi-ECCV-LRR-2016}} & 69.7 & 71.8\\
 \multicolumn{2}{l}{Deeplab v2-CRF~\cite{Chen-Arxiv-Deeplabv2-2016}} & 70.4 & -\\
 \multicolumn{2}{l}{Piecewise~\cite{Lin-CVPR-Piecewisetraining-2016}} & 71.6 & -\\
 \multicolumn{2}{l}{RefineNet~\cite{Lin-CVPR-Refinenet-2017}} & 73.6 & -\\
 \multicolumn{2}{l}{SegModel~\cite{Shen-CVPR-SegModel-2017}} & 78.5 & 79.2 \\
 \multicolumn{2}{l}{DUC~\cite{Wang-CVPR-DUC-2017}} & 77.6 & 80.1 \\
 \multicolumn{2}{l}{PSPNet~\cite{Zhao-CVPR-PSPNet-2017}} & 78.4 & 80.2\\
 \hline
 \noalign{\smallskip}
 \multicolumn{2}{l}{Ours} & \textbf{79.3} & \textbf{80.3}\\ 
 \bottomrule
 \end{tabular}
 \end{center}
 \end{table}

\vspace{-1ex}
\section{Conclusion}
\label{sec:conclusion}
We redefine the semantic segmentation from a macroscopic view of point, regarding it as a task to assign a consistent semantic label to one category of objects, rather than to each single pixel. Inherently, this task requires the intra-class consistency and inter-class distinction. Aiming to consider both sides, we propose a Discriminative Feature Network, which contains two sub-networks: Smooth Network and Border Network. With the bidirectional stage-wise mechanism, our approach can capture the discriminative features for semantic segmentation. Our experimental results show that the proposed approach can significantly improve the performance on the PASCAL VOC 2012 and Cityscapes benchmarks.

\vspace{-1ex}
\section*{Acknowledgment}
This work has been supported by the Project of the National Natural Science Foundation of China No.61433007 and No.61401170.

{\small
\bibliographystyle{ieee}
\bibliography{egbib}
}

\end{document}